\documentclass{article}

\usepackage[preprint]{mtvit}
\usepackage[utf8]{inputenc} 
\usepackage[T1]{fontenc}    
\usepackage{hyperref}       
\usepackage{url}            
\usepackage{booktabs}       
\usepackage{amsfonts}       
\usepackage{nicefrac}       
\usepackage{microtype}      
\usepackage{xcolor}         
\usepackage{graphicx}       
\usepackage{amsmath}        
\usepackage{float} 

\title{Multi-Scale And Token Mergence: Make Your ViT More Efficient}

\author{
Zhe Bian, Zhe Wang, Wenqiang Han, \textbf{Kangping Wang\thanks{Corresponding author.}} \\
Jilin University \\
\texttt{\{bianzhe21,hanwq20\}@mails.jlu.edu.cn, \{wz2000,wangkp\}@jlu.edu.cn}\\
}

\begin{document}

\maketitle

\begin{abstract}
    Since its inception, Vision Transformer (ViT) has emerged as a prevalent model in the computer vision domain. Nonetheless, the multi-head self-attention (MHSA) mechanism in ViT is computationally expensive due to its calculation of relationships among all tokens. Although some techniques mitigate computational overhead by discarding tokens, this also results in the loss of potential information from those tokens. To tackle these issues, we propose a novel token pruning method that retains information from non-crucial tokens by merging them with more crucial tokens, thereby mitigating the impact of pruning on model performance. Crucial and non-crucial tokens are identified by their importance scores and merged based on similarity scores. Furthermore, multi-scale features are exploited to represent images, which are fused prior to token pruning to produce richer feature representations. Importantly, our method can be seamlessly integrated with various ViTs, enhancing their adaptability. Experimental evidence substantiates the efficacy of our approach in reducing the influence of token pruning on model performance. For instance, on the ImageNet dataset, it achieves a remarkable 33\% reduction in computational costs while only incurring a 0.1\% decrease in accuracy on DeiT-S.
\end{abstract}

\section{Introduction}

The transformer architecture~\cite{transformer} has introduced the ability to model global relationships, which was not found in previous convolution-based methods. This has led to impressive performance on a variety of computer vision tasks, including image classification~\cite{vit, deit, t2t}, semantic segmentation~\cite{seg, medical}, object detection~\cite{dert, lrdert}, and image generation~\cite{transgan}. Furthermore, a multitude of supervised, unsupervised, and alternative training techniques~\cite{cdtrans, updert, semi} have been developed for these tasks. Nonetheless, the quadratic computational complexity of the Vision Transformer (ViT), which stems from dense long-range dependencies among image tokens, presents a considerable challenge in computational cost. In general, ViTs require more training iterations and larger datasets than convolutional neural networks (CNNs).

\begin{figure}[t]
    \centering
    \includegraphics[width=0.99\columnwidth]{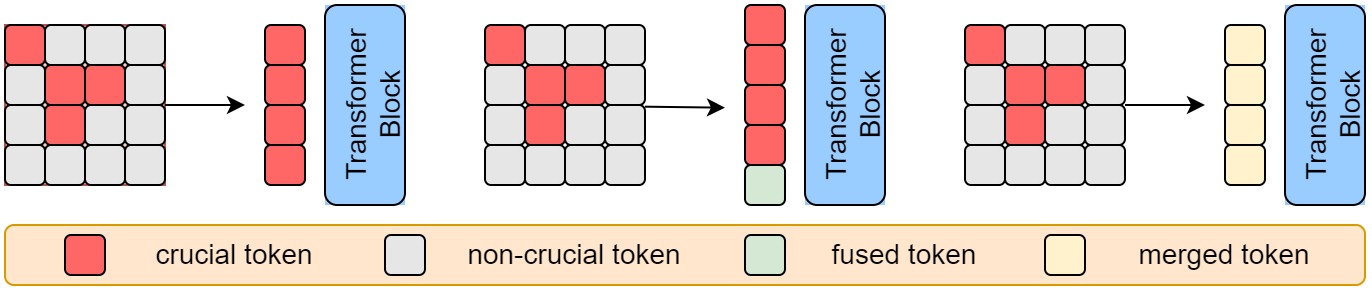}
    \caption{Different handling methods for non-crucial tokens, (a) Only crucial tokens are reserved and non-crucial tokens are discarded. (b) Crucial tokens are kept, and non-crucial tokens are fused into a new token. (c) Non-crucial and crucial tokens are merged.
    }
    \label{fig:contrast}
\end{figure}

To address the computational burden of ViTs, researchers have developed various techniques to reduce internal feature information within the model. One effective strategy involves evaluating token importance and conducting selective pruning. DynamicViT~\cite{dyvit} introduces a module for determining token importance, while EViT~\cite{evit} leverages the class token to assess other tokens' significance. Evo-ViT~\cite{evo-vit} considers the global class token's importance and updates non-crucial tokens differently. AS-ViT~\cite{asvit} employs a learnable threshold to adaptively control the number of retained tokens, and AdaViT~\cite{adavit} introduces modules to evaluate the importance of block, head, and token, streamlining the model from multiple aspects. Although many of these methods directly discard non-crucial tokens, some fuse them based on their importance scores. However, it is important to note that the dropped information may be valuable for classification tasks, as argued in~\cite{nos}. Therefore, it is essential to adopt more effective processing techniques for non-crucial tokens, rather than merely discarding or fusing them. This approach preserves as much information as possible while mitigating the impact of token pruning on model accuracy.

Building on the previous findings, we present a novel approach to token pruning in this paper. Firstly, we assess the relevance of tokens in specific layers. Then, these tokens are classified into two groups: crucial and non-crucial, based on their relative important score. To preserve as much token information as possible while also reducing the number of tokens, we introduce an innovative Token Merging Module. This module calculates similarity metrics between crucial and non-crucial tokens, and merges them based on the metrics. Figure \ref{fig:contrast} illustrates the differences between our method with the previous works.

To improve the accuracy of feature representation, we draw inspiration from the groundbreaking methods introduced in CrossViT~\cite{cvit} and MPViT~\cite{mpvit}. In our approach, we integrate multi-scale features within token embedding to achieve a richer feature set. To counteract potential precision loss during token pruning, we align and fuse multi-scale features, thereby enhancing the overall feature information. However, the incorporation of multi-scale features demands increased computations, which was tackled by an innovative technique. This strategy separates our approach from the previous works in CrossViT~\cite{cvit} and MPViT~\cite{mpvit} too.

We introduce an innovative token pruning technique that achieves a 33\% reduction in FLOPs and a 0.1\% precision increase when applied to DeiT-S~\cite{deit}. Notably, our approach is compatible with the most existing pruning methods. The experiments results show that as the proportion of pruned tokens rises, the module's performance improves significantly. For instance, our technique improves accuracy by 0.2\% when applied to EViT~\cite{evit} with a 0.7 keep rate, and by 0.7\% with a 0.5 keep rate.

In summary, our contributions are twofold:
\begin{itemize}
\item We propose a novel token pruning strategy that effectively utilizes the original input information. Our method strikes a balance between model accuracy and speed, preserving as much original information as feasible during the pruning process. 
\item We improve model accuracy by incorporating multi-scale features and reducing computation prior to token pruning. 
\end{itemize}

\section{Related work}

\paragraph{Vision Transformers.}

Transformer~\cite{transformer} has been successfully applied to natural language processing (NLP) tasks, achieving state-of-the-art results. Furthermore, ViT~\cite{vit} has achieved impressive results by converting images into sequential tokens and processing them through the transformer architecture. Transformers recently have applied to computer vision (CV) domain, producing promising results in image classification~\cite{deit}, object detection~\cite{dert}, semantic segmentation~\cite{segvit}, and other tasks~\cite{transgan}. Simultaneously, several high-performing backbone models have emerged. For example, DeiT~\cite{deit} exploits knowledge distillation to effectively train ViT~\cite{vit} on ImageNet alone, while Swin Transformer~\cite{swin} incorporates local and sliding window operations to introduce inductive bias for ViT~\cite{vit}, consequently reducing its computational complexity. LV-ViT~\cite{lvvit} enhances model accuracy by computing losses for all tokens.

\paragraph{Efficient ViTs.}

Unlike natural language, images often contain a substantial amount of redundant information. As a result, for transformer models with high computational demands, techniques for reducing the computational burden by pruning redundant information from images have become increasingly important. Various approaches for pruning redundant information from images have developed in recent years. One such method focuses on pruning solely the input token information. DynamicViT~\cite{dyvit} assesses the importance of input tokens using a learnable module and determines the number of tokens to retain according to a predefined keep rate. EViT~\cite{evit} and Evo-ViT~\cite{evo-vit} evaluate the importance of other tokens by considering the characteristics of the class token. The former approach merges non-crucial tokens into a new token, while the latter retains non-crucial tokens while employing rapid updates to maintain information flow integrity. A-ViT~\cite{avit} dynamically adjusts the number of tokens according to the input's complexity, thereby controlling the model's computational complexity. AS-ViT~\cite{asvit} emphasizes the role of distinct heads and employs a learnable threshold to determine token retention. Another method involves pruning token, head, block, and other model components by analyzing redundancy from multiple perspectives. AdaViT~\cite{adavit} introduces learnable parameters to determine whether block, head, and token should be pruned. MIA-Former~\cite{miavit} introduces an MIA-Controller to decide if a block should be skipped; if not, a learnable module is employed to determine head and token pruning.

\section{Method}
\subsection{Preliminaries}

Drawing inspiration from the successful implementation of Transformer~\cite{transformer} in the natural language processing domain, Vision Transformer (ViT)~\cite{vit} converts an image into a sequence of tokens, analogous to words in a sentence. It also incorporates a class token (CLS) to obtain an image representation. To differentiate between tokens at various positions, position embeddings are integrated. The resulting input is then fed into stacked transformer encoders, with the output features forming the foundation for downstream tasks. The ViT's core component, the transformer encoder, comprises multi-head self-attention (MHSA) and a feed-forward network (FFN). MHSA is employed to obtain the query, key, and value through a linear mapping of the input information. Subsequently, the query and key are multiplied to produce an attention map, which is ultimately multiplied by the value and outputted via a matrix mapping. For an input $X\in {{R}^{N \times D}}$ to ViT, the MHSA process can be mathematically expressed as follows:
\begin{equation}
    \begin{split}
       &{{Q}_{i}}={{X}_{i}}W_{Q}^{i};{{K}_{i}}={{X}_{i}}W_{K}^{i} ;{{V}_{i}}={{X}_{i}}W_{V}^{i},\\
       &{\rm Attention}({{Q}_{i}},{{K}_{i}},{{V}_{i}})={\rm Softmax}(\frac{{{{Q}}_{i}}{K}_{i}^{T}}{\sqrt{d}}){{V}_{i}},\\
       &{\rm MHSA}(X)={{W}_{O}}{\rm concat}[{\rm Attention}({{Q}_{i}},{{K}_{i}},{{V}_{i}})]_{i=1}^{H},\\   
 \end{split}
\end{equation}
Let $i\in \{1,2,\ldots {,H}\}$ denote the index of the i-th head in a multi-head attention mechanism. Let ${{X}_{i}},{{Q}_{i}},{{K}_{i}},{{V}_{i}}\in {{R}^{N \times d}}$ be the input, query, key, and value matrices of the i-th head, where $N$ is the number of tokens, $D$ is the embedding dimension, and $d$ is the dimension of the embedding of a single head.
\begin{equation}
    \begin{split}
{\rm FFN}(X)={\rm Sigmoid}({\rm Linear}({\rm GeLU}({\rm Linear}(X)))).
 \end{split}
\end{equation}
\subsubsection{Computation complexity}
In the ViT architecture, the computational cost of the MHSA and FFN modules are given by $O(4N{{D}^{2}}+2{{N}^{2}}D)$ and $O(8N{{D}^{2}})$, respectively. Therefore, the total computational cost of a transformer block can be estimated as $O(12N{{D}^{2}}+2{{N}^{2}}D)$. Token pruning methods can be used to reduce the number of tokens by a certain percentage, denoted by $\lambda \%$. As a result, such methods can effectively reduce the FLOPs of a transformer block by at least $\lambda \%$.

\subsection{Token selection}

In the context of image classification, ViT solely relies on the class token for classification-related information. Specifically, we employ the following formulation \ref{eq:cls}:
\begin{equation}
    \begin{split}
{{X}_{CLS}}={\rm Softmax}(\frac{{{Q}_{CLS}}{{K}^{T}}}{\sqrt{d}})V.
\label{eq:cls}
     \end{split}
\end{equation}
where ${{Q}_{CLS}}$ represents the query vector's class token. The class token ${X}_{CLS}$ constitutes a linear combination of all token values, enabling its attention score to encapsulate the relative significance of other tokens for classification outcomes. In alignment with previous methods~\cite{dyvit, evit, evo-vit}, we merely average dimensions within the head. During token selection, we preserve the first N tokens possessing the highest attention scores as crucial tokens, while the remaining tokens are classified as non-crucial tokens; the class token is inherently considered an crucial token by default.

\subsection{Token mergence}

\begin{figure}[t]
    \centering
    \includegraphics[width=0.99\columnwidth]{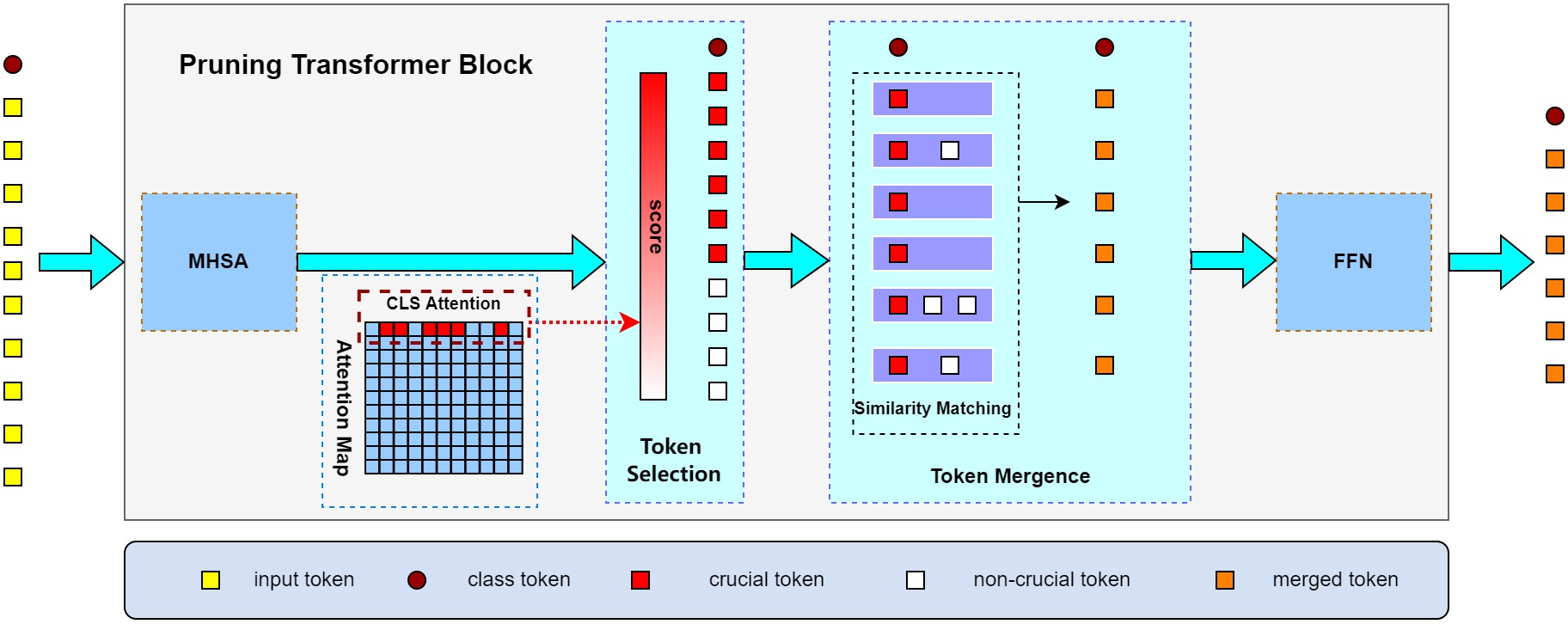}
    \caption{Token pruning within a single transformer encoder. Like EViT, We use the value of CLS attention as the score to judge the importance of each token. And identify the top-k crucial tokens. Then, we use cosine similarity of crucial and non-crucial tokens as similarity evaluation criterion and merge them. }
    \label{fig:idea}
\end{figure}

Most token pruning techniques depend on a predefined criterion to decide whether to retain or prune a token. However, this reduction of input data for subsequent transformer layers can lead to significant accuracy losses when many tokens are pruned. To address this issue, our approach aims to prioritize crucial token information processing while concurrently managing non-crucial token information. We propose merging non-crucial token information into the most similar crucial token, as depicted in Figure \ref{fig:idea}.

In our method, tokens are classified into two categories: crucial and non-crucial. To prevent the pruning of non-crucial tokens from negatively impacting the model's accuracy, A merging stage is introduced for these two categories before pruning. Cosine similarity is employed as the primary metric for calculating the similarity between the categories, chosen for its straightforward implementation. We determine the similarity between each non-crucial token and the set of crucial tokens by computing the cosine similarity of the non-crucial token with each token in the crucial set. Subsequently, for each non-crucial token, we identify the crucial token with the highest cosine similarity as its most similar counterpart and merge the tokens. The weighted merging method is utilized to accommodate the varying relevance of tokens. The weights are determined by corresponding item in class token. Formula \ref{eq:fuse} represents the merged token. 
\begin{equation}
    \begin{split}
{{m}_{j}}={({{w}_{j}}{{i}_{j}}+\sum\limits_{k\in {{S}_{j}}}{{{w}_{k}}{{u}_{k}}})}/{({{w}_{j}}+\sum\limits_{k\in {{S}_{j}}}{{{w}_{k}}})}.\;
\label{eq:fuse}
 \end{split}
\end{equation}
Where ${{w}_{j}}$ denotes the importance score of the crucial token ${{i}_{j}}$, ${{w}_{k}}$ denotes the importance score of the non-crucial token ${{u}_{k}}$ and ${{S}_{j}}$ denotes the set of non-crucial tokens merged with the j-th crucial token, $m_j$ denotes the j-th merged token. The size of merged tokens is calculated from keep rate. 

\subsection{Multi-scale features}

To achieve higher accuracy, input tokens should present as many features as possible before token pruning. A multi-scale feature extraction approach is proposed for obtaining richer feature representations, with the complete model architecture illustrated in Figure \ref{fig:framework}. Specifically, we generate two feature groups with identical embedding dimensions but varying token counts during feature embedding. The group with more tokens is termed as high-scale features, while the group with fewer tokens is termed low-scale features. We assign learnable position codes to each feature group. Before token pruning stage, we fuse the information from both feature groups to obtain more representative features. During the multi-scale feature fusion process, we upsample low-scale features using nearest interpolation and convolution methods to align them with high-scale features in terms of token count. To reduce parameter count and ensure effectiveness, we employ the LKA module~\cite{van} to transform the upsampled features. Finally, we add the upsampled low-scale features and high-scale features in the token dimension and integrate new location information into the resulting features using the PEG module~\cite{peg}. The formulaic description of this process is as follows:
\begin{figure}[t]
    \centering
    \includegraphics[width=0.99\columnwidth]{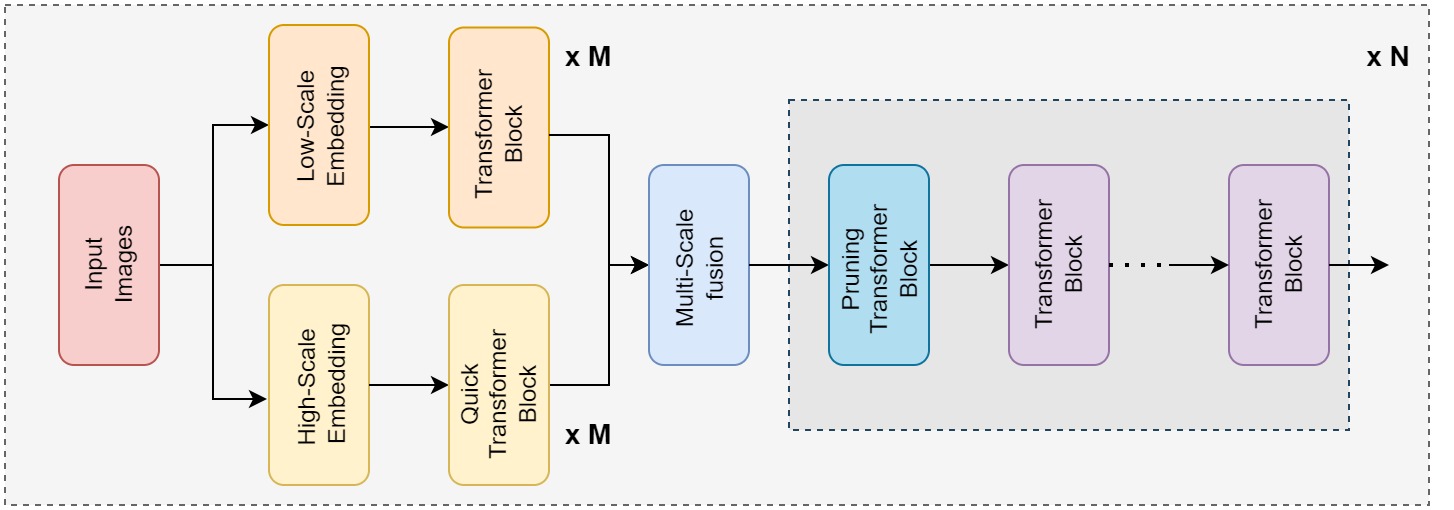}
    \caption{The overall architecture of the multi-scale transformer model.}
    \label{fig:framework}
\end{figure}
\begin{equation}
    \begin{split}
 X={\rm PEG}({X}_{h}+{\rm LKA}({\rm UP}({X}_{l}))).
 \end{split}
\end{equation}
Where ${X}_{h}$ and ${X}_{l}$ respectively represent the high-scale features and the low-scale features, UP represents the nearest interpolation operation.

\subsubsection{Reduce computation}
Incorporating multi-scale features in visual transformers unavoidably increases computational demands. To mitigate this overhead while maintaining high accuracy, we propose a simple modification to the attention modules in the first blocks of the high-scale feature hierarchy. Specifically, we first downsample the high-scale features before processing them with the MHSA module. The downsampled tokens are then input into the MHSA module, and the resulting feature vectors are upsampled to the original scale, keeping the class token unchanged during the whole of the process. The formulaic description of this process is as follows:
\begin{equation}
    \begin{split}
 {X}_{h}={X}_{h} + {\rm UP}({\rm MHSA}({\rm DOWN}({X}_{h}))).
 \end{split}
\end{equation}
Where ${X}_{h}$ denotes the high-scale feature, DOWN represents a simple downsample operation, and UP refers to a nearest interpolation method. This update ensures that each block's multi-scale feature computational requirements are approximately equivalent to those of the original ViT.

\section{Experiments}

\subsection{Implementation details}
To maintain consistency, we trained all models in our experiment on the ImageNet-1k dataset~\cite{imagenet}, containing a training set of 12 million images and a test set of 50k images. To enhance performance, we incorporate Token Selection and Token Mergence modules into the ${4}^{th}$, ${7}^{th}$, and ${10}^{th}$ layers of DeiT-T, DeiT-S, and DeiT-B~\cite{deit}, as well as into the ${5}^{th}$, ${9}^{th}$, and ${13}^{th}$ layers of LV-ViT-S~\cite{lvvit}, adopting the same optimization strategy as the original paper. We employed high-scale features with 196 tokens and low-scale features with 49 tokens, each containing a class token. For a fair comparison with EViT~\cite{evit}, we gradually reduced the keep rate of attentive tokens from 1 to the target value utilizing a cosine schedule. Following the other methods' setting, we trained all models for 300 epochs on 8 NVIDIA RTX 3090 GPUs, and measured their throughput using a single NVIDIA RTX 3090 GPU with a 128 batch size.

\subsection{Main results}

\subsubsection{Comparison with the-state-of-the-arts}
As shown in Table \ref{tab:deits}, we provide a comparative analysis of our token pruning method against other state-of-the-art techniques on DeiT~\cite{deit}, reporting top-1 accuracy and FLOPs to assess performance. Our approach outperforms previous methods, delivering superior results while maintaining reasonable computational costs. Specifically, our method reduces the computational complexity of DeiT-T~\cite{deit} by 35\% while improving model accuracy by 0.5\%. We also apply our method to LV-ViT~\cite{lvvit}, which is a deep-narrow architecture. As illustrated in Table \ref{tab:lvvits}, our approach enables LV-ViT~\cite{lvvit} to achieve an optimal balance between accuracy and speed compared to other transformer models.
 \begin{table}[t]
    \caption{Comparison with existing token pruning methods.}
    \label{tab:deits}
    \centering
    \resizebox{1.0\columnwidth}{!}{

        \begin{tabular}[t]{l|ccc|c}
            \toprule
            Model   & Params(M) & FLOPs(G) & FLOPs$\downarrow$(\%) & Top-1(\%)   \\
            \midrule
            DeiT-T~\cite{deit}       & 5.7  & 1.3  & 0.0  & 72.2          \\
            \midrule
            DynamicViT~\cite{dyvit}  & 5.9  & 0.9  & 30.8 & 71.2(-1.0)            \\
            SP-ViT~\cite{spvit}      & 5.7  & 0.9  & 30.8 & 72.1(-0.1)            \\
            Evo-ViT~\cite{evo-vit}   & 5.7  & 0.8  & 38.5 & 72.0(-0.2)            \\
            \textbf{Ours-DeiT-T}     & 5.8  & 0.8  & 38.5 & \textbf{72.7(+0.5)}            \\
            \midrule
            DeiT-S~\cite{deit}       & 22.1  & 4.6  & 0.0  & 79.8          \\
            \midrule
            ToMe~\cite{tome}         & 22.1  & 2.7  & 41.3 & 79.4(-0.4)           \\
            DynamicViT~\cite{dyvit}  & 22.8  & 2.9  & 37.0 & 79.3(-0.5)           \\
            A-ViT~\cite{avit}        & 22.1  & 3.6  & 21.7 & 78.6(-1.2)           \\
            Evo-ViT~\cite{evo-vit}   & 22.1  & 3.0  & 34.8 & 79.4(-0.4)           \\
            EViT~\cite{evit}         & 22.1  & 3.0  & 34.8 & 79.5(-0.3)           \\
            AS-ViT~\cite{asvit}      & 22.1  & 3.0  & 34.8 & 79.6(-0.2)           \\
            \textbf{Ours-DeiT-S}     & 22.2  & 3.1  & 32.6 & \textbf{79.7(-0.1)}           \\
            \midrule
            DeiT-B~\cite{deit}       & 86.6  & 17.5  & 0.0  & 81.8          \\
            \midrule
            DynamicViT~\cite{dyvit}  & -     & 11.2  & 36.0 & 81.3(-0.5)          \\
            Evo-ViT~\cite{evo-vit}   & 86.6  & 11.5  & 34.2 & 81.3(-0.5)         \\
            EViT~\cite{evit}         & 86.6  & 11.5  & 34.2 & 81.3(-0.5)         \\
            AS-ViT~\cite{asvit}      & 88.6  & 11.2  & 36.0 & 81.4(-0.4)          \\
            \textbf{Ours-DeiT-B}     & 86.6  & 11.5  & 34.2 & \textbf{81.5(-0.3)}         \\
            \bottomrule
        \end{tabular}
    }
    
\end{table}

\begin{table}[t]
    \caption{Comparison with the state-of-the-art models.}
    \label{tab:lvvits}
    \centering
    \resizebox{1.0\columnwidth}{!}{

        \begin{tabular}[t]{lccc}
            \toprule
            Method      & Params(M) & FLOPs(G) & Top-1(\%)     \\
            \midrule
            ViT-B~\cite{vit}         & 86.6    & 4.6  & 77.9   \\
            DeiT-S~\cite{deit}       & 29.0    & 4.5  & 79.8   \\
            Swin-T~\cite{swin}       & 50.0    & 4.6  & 81.3   \\
            T2T-ViT-14~\cite{t2t}    & 21.5    & 4.8  & 81.5   \\
            CvT-21~\cite{cvt}        & 31.5    & 7.1  & 82.5   \\
            DW-T~\cite{dw}           & 30.0    & 5.2  & 82.0   \\
            Cross-ViT-S~\cite{cvit}   & 26.7    & 5.6  & 81.0   \\
            CoaT-Lite Small~\cite{coat} & 20.0  & 4.0  & 81.9   \\
            RegionViT-S~\cite{reginvit} & 30.3  & 5.3  & 82.6  \\
            LV-ViT-S~\cite{lvvit}    & 26.2    & 6.6  & 83.3   \\
            \midrule
            DynamicViT-LV-S~\cite{dyvit}   & 26.9    & 3.7 & 82.0   \\
            EViT-LV-S~\cite{evit}         & 26.2    & 3.9 & 82.5   \\
            AS-LV-S~\cite{asvit}          & 26.2    & 3.9 & 82.6   \\
            \textbf{Ours-LV-S}            & 26.2    & 3.9 & \textbf{82.8}  \\
            \bottomrule
        \end{tabular}
    }
    
\end{table}

\subsubsection{Comparison with existing methods on each keep rate}
\begin{figure}[htbp]
    \begin{minipage}{0.49\linewidth}  
    \centering
    \includegraphics[width=0.99\columnwidth]{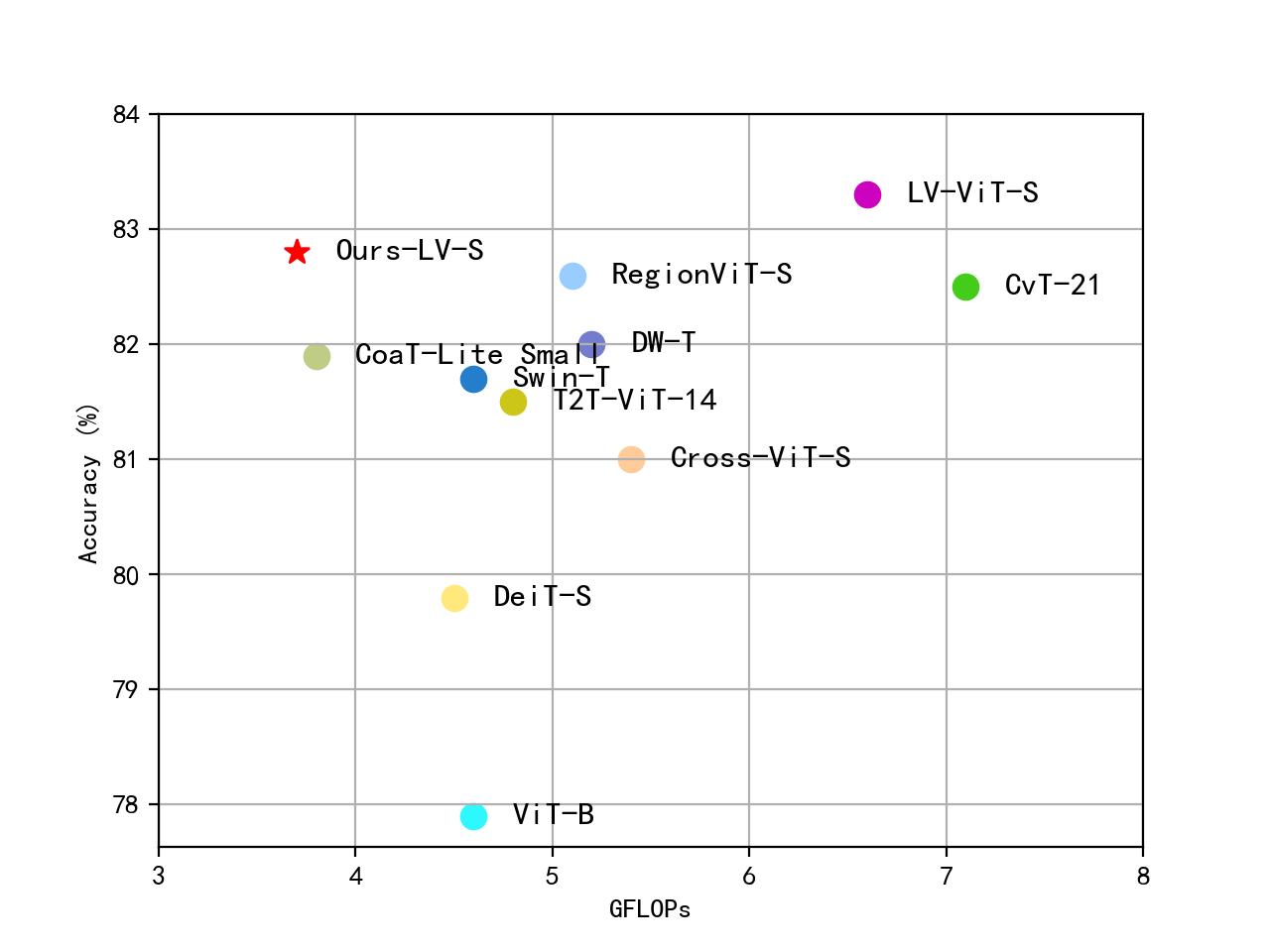}
    {\caption{Comparison with the state-of-the-art models.}
    \label{fig:sota}}
    \end{minipage}
    \begin{minipage}{0.49\linewidth}
    \centering
    \includegraphics[width=0.99\columnwidth]{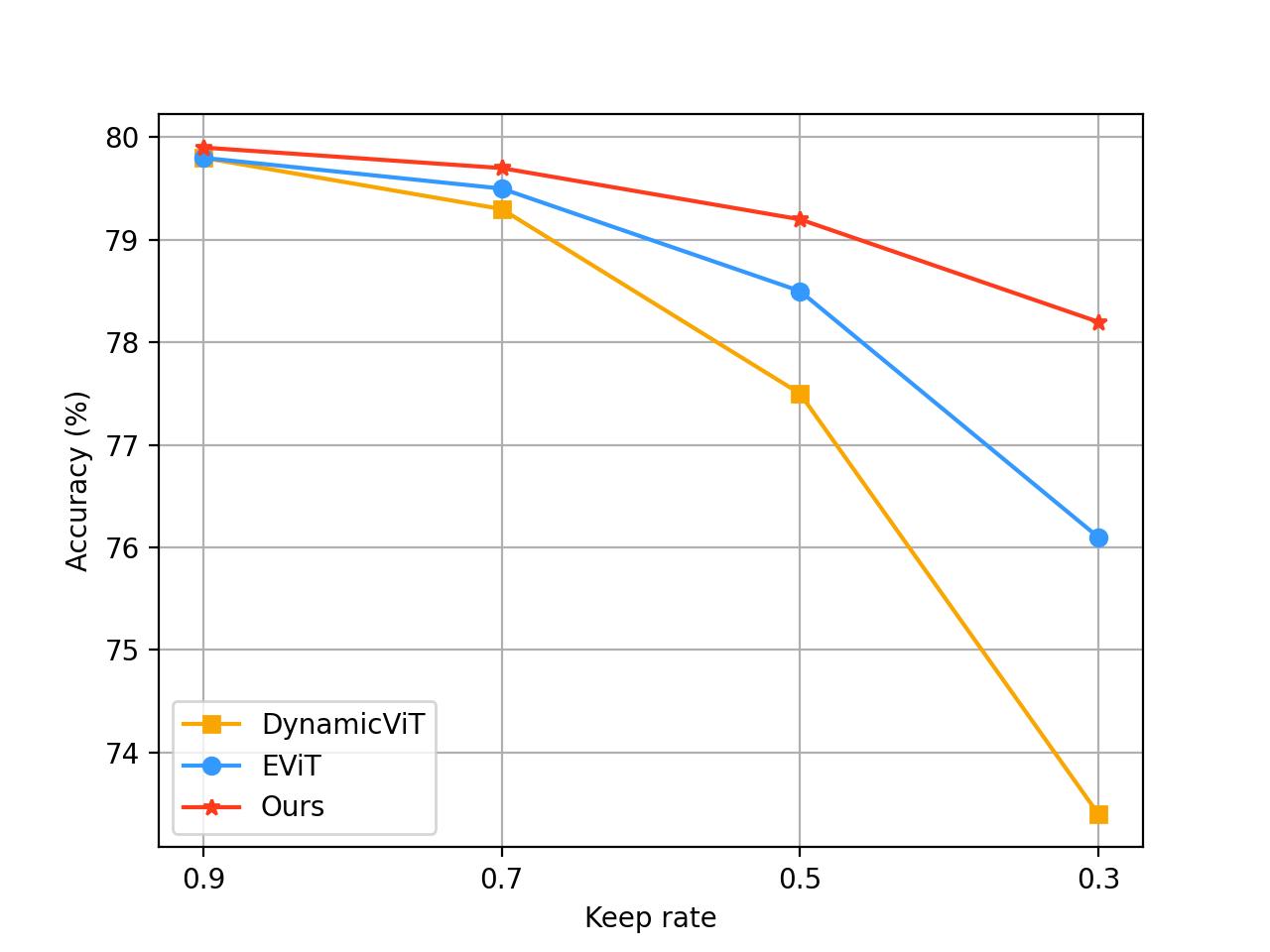}
    {\caption{Comparisons with existing methods on each keep rate.}
     \label{fig:ekr}}
     \end{minipage}
\end{figure}

We evaluate the accuracy of our method and the other two \cite{dyvit, evit} at different keep rates, as depicted in Figure \ref{fig:ekr}. The results show that our method outperforms the other methods at the same keep rates, with more significant improvements at lower keep rates. This is likely because our method merges the information of non-crucial tokens into crucial tokens, reducing information loss. This is in contrast to Dynamic-ViT~\cite{dyvit}, which directly discards non-crucial tokens, or EViT~\cite{evit}, which fuses them into a single token. Consequently, the lower the keep rate, the more tokens are merged and the greater the benefit of our method. Furthermore, we incorporate multi-scale features before token pruning, leading to improved performance compared to the baseline at high keep rates.

\subsubsection{Visualization}

\begin{figure*}[t]
    \centering
    \includegraphics[width=\textwidth]{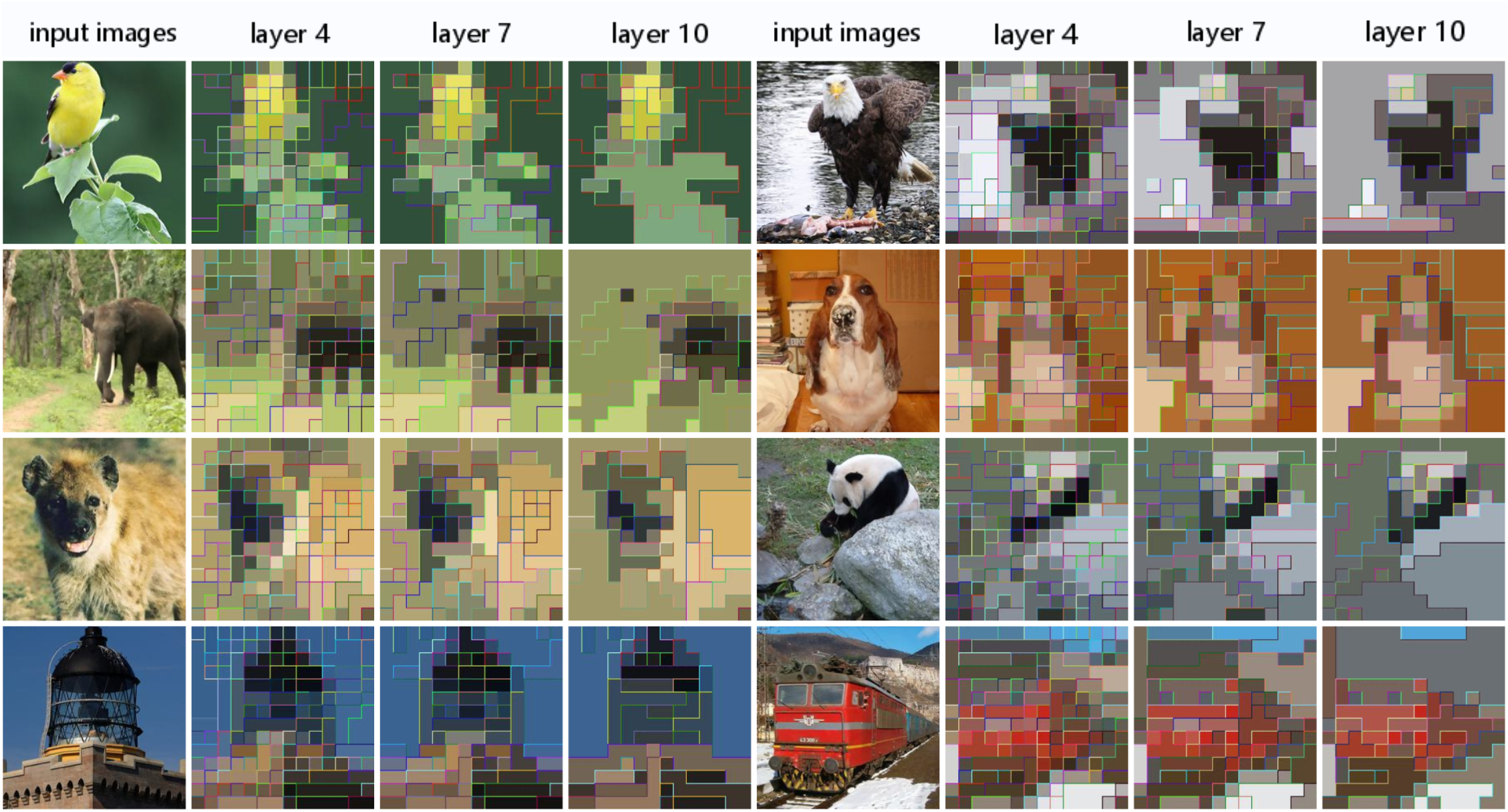}
    \caption{Visualization of the token pruning process on DeiT-S at a keep rate of 0.5. Using a color-coding scheme, tokens that have been merged together are represented with the same color, allowing for a clear and intuitive representation of the pruning process.}
    \label{fig:vis}
\end{figure*}

Our method adopts a novel strategy to decrease computational costs by identifying and preserving crucial tokens while merging non-crucial ones to reduce redundancy. To evaluate the effectiveness of our approach, we present visualizations of each stage in Figure \ref{fig:vis}. The results show that our method successfully distinguishes crucial tokens from non-crucial ones, by concentrating tokens in positions that significantly impact classification, while merging background elements. For instance, in the image classified as a panda, our method assigns more tokens to the panda after each pruning layer, while continuously merging the background. This evidence depicts why the method is effective.

\subsection{Ablation Analysis}
\subsubsection{Effectiveness of each module}
We have integrated our approach into EViT~\cite{evit} to evaluate the effectiveness of individual modules. The experimental results are shown in Table \ref{tab:module}, where TM denotes token mergence and MF represents multi-scale feature fusion. Our analysis shows that incorporating TM significantly improves the baseline accuracy by preserving non-crucial token feature information. Furthermore, the inclusion of MF results in more comprehensive feature extraction and improvement in baseline accuracy.
\begin{table}[t]
\begin{minipage}[t]{0.48\linewidth}
    \makeatletter\def\@captype{table}
    \caption{The effectiveness of each module on EViT-DeiT-S with different keep rates.}
    \label{tab:module}
    \begin{tabular}{lll}
         \toprule
        Method                & Top-1 (\%) & FLOPs(G) \\
        \hline
        \multicolumn{3}{c}{DeiT-S/$\eta$=0.5}\\       
        \hline
        baseline               &  78.5      & 2.3  \\
        +TM                    &  79.1      & 2.3  \\
        +MF                    &  79.2      & 2.4 \\
        \hline
       \multicolumn{3}{c}{DeiT-S/$\eta$=0.7}\\                         
        \hline
        baseline               &  79.5     & 3.0  \\
        +TM                    &  79.6     & 3.0  \\
        +MF                    &  79.7     & 3.1  \\
        \hline
    \end{tabular}
    
\end{minipage}
\begin{minipage}[t]{0.48\linewidth}  
    \makeatletter\def\@captype{table}
    \caption{Different similarity calculation methods on DeiT-S.}
    \label{tab:similarity}
    \begin{tabular}{lll}
        \toprule
        Method                & Top-1 (\%) & Throughput (img/s) \\
        \hline
        Random                &   78.8         &  2137  \\
        Attention Map         &   79.0         &  2195  \\
        L1 Distance           &   79.1         &  2033  \\
        L2 Distance           &   79.1         &  2181  \\ 
        Ours                  &   79.2         &  2201  \\
        \hline
    \end{tabular}
    
  \end{minipage}
\end{table}

\subsubsection{Different similarity calculation methods}
As shown in Table \ref{tab:similarity}, we have compared several techniques for computing the similarity between crucial and non-crucial tokens on DeiT-S~\cite{deit} when the keep rate is 0.5. These methods include: \romannumeral1) merging crucial and non-crucial tokens randomly; \romannumeral2) assessing similarity using the cross scores of crucial and non-crucial tokens in the attention matrix; \romannumeral3) employing Manhattan distance for similarity calculation; \romannumeral4) employing Euclidean distance for similarity computation; and \romannumeral5) employing cosine similarity. Our observations indicate that, given similar throughput, our method typically attains a 0.1\% enhancement in accuracy relative to alternative similarity calculation techniques. This implies that cosine similarity serves as an effective metric for determining similarity between crucial and non-crucial tokens. Moreover, compared to other methods, cosine similarity has the relatively simple computation. In summary, our findings endorse the application of cosine similarity for computing similarity between crucial and non-crucial tokens, owing to its precision and computational efficiency.

\section{Conclusion}

In this study, we introduce an innovative token pruning module that is designed to reduce the impact on model accuracy during token pruning. By selectively identifying crucial tokens and merging non-crucial ones, we maintain the accuracy caused by token pruning. The integration of multi-scale features further increases model accuracy. The experiments results from DeiT~\cite{deit} support the effectiveness of our proposed module. Moreover, our technique can be combined with the majority of existing token pruning methods to enhance their accuracy with almost no computational overhead. Our approach achieves an good balance between accuracy and speed. We envision our method being valuable for downstream tasks such as target detection, semantic segmentation, and beyond classification tasks.

{\small

\bibliographystyle{plain}
}

\end{document}